\documentclass[conference]{IEEEtran}
\IEEEoverridecommandlockouts
\usepackage{cite}
\usepackage{amsmath,amssymb,amsfonts}
\usepackage{algorithmic}
\usepackage{graphicx}
\usepackage{textcomp}
\usepackage{xcolor}
\def\BibTeX{{\rm B\kern-.05em{\sc i\kern-.025em b}\kern-.08em
    T\kern-.1667em\lower.7ex\hbox{E}\kern-.125emX}}

\usepackage{multirow}
\usepackage{tikz}
\usepackage{color}
\usepackage{pifont}
\usepackage{pgfplots}
\usepgfplotslibrary{groupplots}
\usepackage{pgfplotstable}
\usepackage{filecontents}

\pgfplotsset{compat=1.18} 

\usepackage{xspace}
\makeatletter
\DeclareRobustCommand\onedot{\futurelet\@let@token\@onedot}
\def\@onedot{\ifx\@let@token.\else.\null\fi\xspace}

\def\ie{\emph{i.e}\onedot} 
 
\def\etc{\emph{etc}\onedot} 
 
\def\etal{\emph{et al}\onedot}
\makeatother

\begin{document}

\title{Improving classification of road surface conditions via road area extraction and contrastive learning
}

\author{Linh Trinh \textit{Student Member IEEE}, Ali Anwar \textit{Member IEEE}, Siegfried Mercelis\\
{Faculty of Applied Engineering, IDLab, University of Antwerp-imec, Belgium} \\
{\tt\small\{linh.trinh, ali.anwar, siegfried.mercelis\}@uantwerpen.be}\\
}

\maketitle

\begin{abstract}
Maintaining roads is crucial to economic growth and citizen well-being because roads are a vital means of transportation. In various countries, the inspection of road surfaces is still done manually, however, to automate it, research interest is now focused on detecting the road surface defects via the visual data. While, previous research has been focused on deep learning methods which tend to process the entire image and leads to heavy computational cost. In this study, we focus our attention on improving the classification performance while keeping the computational cost of our solution low. Instead of processing the whole image, we introduce a segmentation model to only focus the downstream classification model to the road surface in the image. Furthermore, we employ contrastive learning during model training to improve the road surface condition classification.  Our experiments on the public RTK dataset demonstrate a significant improvement in our proposed method when compared to previous works. 
\end{abstract}


\section{Introduction}
One of the most important parts of transportation is roads. The issue of inspecting the condition of road surfaces for maintenance purposes is a topic of research in several fields, including autonomous driving and road engineering. Automatic road defects measurements are necessary for efficient and cost-effective road maintenance because cities and municipalities invest substantial funds into fixing damaged components. There are several standards for visual surface assessment, including the Pavement Condition Index (PCI) \cite{astm2012standard}, Pavement Surface Condition Index (PSCI) \cite{psci}, and Pavement Surface Evaluation Rating (PASER) \cite{paser}. For the purpose of conducting pavement evaluations, these standards offer measurement instructions for pavement characteristics. Nowadays, thanks to advancements in AI and deep learning solutions, road data can be used to automate maintenance tasks with a high degree of accuracy. To be more precise, a vehicle equipped with sensors like cameras, lidar, radar, or GPS can be driven along a road to gather data. Then, a machine learning model can be used to assess the state of the road. A big data-driven approach to their construction has been heavily emphasized in relation to these tasks. 
More recently, a number of studies involving datasets and methodologies from different nations have been publicized, such as LTPP-FHWA in the USA \cite{data_ltpp}, GAPs in Germany \cite{data_GAPs}, CFD from China \cite{cfd_china}, and RDD in Japan, India, the Czech Republic, and Norway \cite{RDD19,RDD20,RDD22}, offering visual data for the evaluation of road surface conditions through the use of artificial intelligence and deep learning techniques. That it is a fascinating area for transportation research is strongly supported by this.

Apart from road maintenance, the road surface is an important object to study in autonomous driving. With the increasing collection of public road data \cite{pp4av,fisheyepp4av} by Original Equipment Manufacturer (OEM), it is crucial to identify road conditions for the purpose of enabling self-driving features \cite{odd@sae,nhtsa_fw} or conducting safety validation \cite{dsa4av}.
For example, the appearance of a dirt road surface may indicate the presence of an ADAS feature that may be or may not be intended to operate on this road. The first step in deciding whether or not to enable or disable particular ADAS features is to detect the type and quality of the road surface.
To summarize, classification of road surface types, as well as their condition, is an important task for both ADAS and Autonomous Navigation Systems (ANS), as well as Road Infrastructure Maintenance Departments conducting regular road maintenance checks. In the context of ADAS or ANS, surface type and quality classification may indicate a safer and more comfortable driving style. In the context of road maintenance, it can speed up and improve the identification of critical points process by automatically identifying portions that require more attention and maintenance. Therefore, developing an efficient method for identifying road surface and road condition could assist with a variety of applications. 

In the field of computer vision, research into road surface and road condition inspection has been on the rise in the past few years. In the realm of road surface classification, numerous new methods have been suggested. Convolutional Neural Networks (CNN) architecture is the backbone of most proposed methods, as exemplified by \cite{road_surface_CNN,road_surface_RCNet,road_surface_condition,road_surface_cnn2,road_surface_dcnn}. These techniques propose the layout by way of neural network structure design or modification. Alternatively, methods such as ResNet50, InceptionNetV3 \cite{nolte2018assessment}, or ConvNeXt \cite{road_surface_fusion} make use of a pretrained backbone on the ImageNet dataset for classification. One alternative is to combine CNN with other types of neural networks, like Transformer \cite{road_surface_cnn-transformer} and Long Short-Term Memory \cite{road_surface_iv} (LSTM). Studies have shown that data fusion and feature fusion are alternative approaches that can improve the classification task, such as \cite{road_surface_fusion, road_surface_shadow,road_surface_cnn-transformer}. A number of methods have been shown in the literature such as \cite{rtk_road_quality,method_bleeding,method_cnn_soict,method_crack_pave_vn} to use Convolutional Neural Networks (CNN) as classifiers for road condition classification. There has been a lot of study on building classifiers using pretrained backbone networks; notable examples include VGG16 and MobileNetV2 \cite{method_cnn_2017,method_transfer_learning,method_light_weight_nn}, ResNet18 \cite{method_raveling}, and ResNet101 \cite{method_orthoframes}. Other initiatives sought to enhance road surface condition inspections by integrating supplementary signals, such as wheel-road interactions \cite{method_embedded}, accelerometers \cite{method_sensor}, orthoframes \cite{method_orthoframes}.

However, existing road condition classification methods require model training on image that include other aspects of the road area such as buildings and traffic participants (\textit{e.g.} car, bus, truck, \etc,), which may not be relevant information for the supporting task. Intuitively, we only need the road area to determine the type and condition of road surface. Work \cite{rtk_road_quality} proposes cropping an image horizontally that contains more road area than others for classification. However, this type of cutting still leaves objects other than the road area. 
In this paper, we present an approach to improve the performance of deep learning driven road surface condition classification. To elaborate, our method consists of two stages: the initial phase involves extracting the road area from image captured by vehicles, while the second phase involves the execution of classification tasks utilizing the extracted road area. Furthermore, we propose combining contrastive learning with the training model to enhance classification tasks because the semantic embedding features extracted from the classification model during learning can be heterogeneous due to a lack of consistency within a class. The remaining sections of the paper are structured as follows. We present our literature studies on the related works in Section \ref{sec:related_work}. Next, in Section \ref{sec:method}, we outline our methodology. Section \ref{sec:exp} provides an analysis of our experiment and the outcomes we obtained. Ultimately, we bring our work to a close in Section \ref{sec:conclude}.

\section{Related works} \label{sec:related_work}
There have been various previous studies which focus on the deep learning driven methodologies for the classification of road surface conditions.
Shi \etal \cite{road_surface_cnn-transformer} propose a CNN-transformer architecture for visual-tactile fusion road recognition using lidar and camera data. The method is tested on private data sets that include four types of roads: marble, painted, gravel, and asphalt.
Pereira \etal \cite{road_surface_CNN} propose a CNN-based deep learning classification method for distinguishing between paved and unpaved roads using images.
Dewangan \etal \cite{road_surface_RCNet} propose a CNN-based road classification network, RCNet, to identify five categories of road surface: curvy, dry, ice, rough, and wet roads.
Cheng \etal \cite{road_surface_condition} propose a deep learning method for road surface condition classifications based on the CNN model. The experiment is being conducted on six different types of roads: dry, wet, snowy, muddy, and other roads.
Chhabra \etal \cite{road_surface_iv} use CNN, LSTM, and InceptionV3 to analyze road surfaces of three types: asphalt, gravel, and concrete.
While, Balcerek \etal \cite{road_surface_cnn2} propose a CNN architecture for classifying road surfaces into different material types such as asphalt, concrete, trylinka, concrete paver blocks, granite paver blocks, openwork surface, gravel, sand, and grass.
Furthermore, Zhang \etal \cite{road_surface_fusion} proposed a multi-supervised bidirectional fusion network model for detecting road surface conditions. The model uses backbone ConvNeXt for feature extraction and a bidirectional fusion module to fuse features and improve classification task.
Zhao \etal \cite{road_surface_data1} provide a dataset for road surface classification collected in China, containing 13 types of road surfaces. A baseline model for classification based on CNN is also presented.
Kim \etal \cite{road_surface_dcnn} developed a deep convolutional neural network to classify eight types of road surfaces: asphalt, cement, dirt, marble, paint, snow, water, and ice.
Nolte \etal \cite{nolte2018assessment} use two CNN networks, ResNet50 and InceptionNetV3, to classify six types of road surfaces.
Tian \etal \cite{road_surface_shadow} propose a framework to remove the shadow from the road area, classify candidate regions of interest with multiple road surface classes, and finally fuse the identification from multiple sensors and Zhao \etal \cite{ZHAO2022108483} created a road surface image dataset that captures diverse road and weather conditions in China. 

\begin{figure*}[ht]
    \centering
    \includegraphics[width=0.89\textwidth]{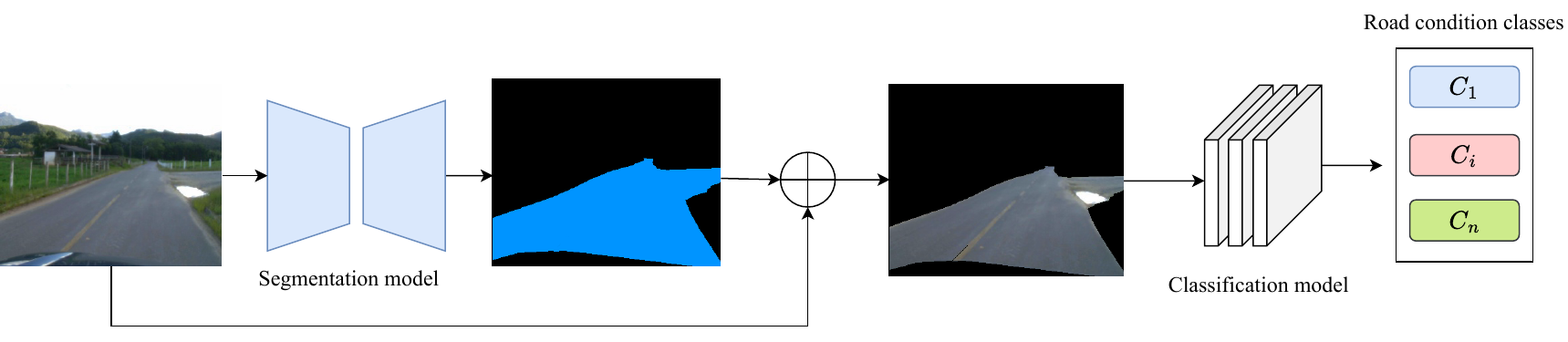}
    \caption{Our proposed framework consists of two stages: road area extraction and classification of road surface conditions.}
    \label{fig:our_fw}
\end{figure*}

Additionally, other studies have confirmed the findings by conducting research on privately curated datasets. For example, Gagliardi \etal \cite{method_embedded} propose a technique to assess the quality of road pavement by analyzing acoustic data from the interaction between wheels and the road. This model has been evaluated using a privately gathered dataset consisting of four distinct categories: silence, unknown, good quality, and ruined. Rateke \etal \cite{rtk_road_quality} propose a CNN architecture for classifying images according to the condition of the road surface. Hasanaath \etal \cite{method_transfer_learning} propose a transfer learning approach that involves fine-tuning a classification model using a pre-trained backbone, such as VGG16 or MobileNetV2. The model underwent testing using a proprietary dataset obtained from the Kingdom of Saudi Arabia. The dataset consisted of four classifications: good, medium, bad, and unpaved. Chaudhary \etal \cite{method_light_weight_nn} propose a classification model that utilizes a pre-trained MobileNetV2 as its foundation, with additional layers added for enhanced performance. The evaluation of an RTK dataset involves three primary classifications: good, normal, and poor quality. Andrades \etal \cite{method_sensor} propose a classification model utilizing the self-organizing map (SOM) algorithm. This technique, which underwent testing using confidential data, comprises multiple sensors, such as an accelerometer and a microcontroller. Riid \etal \cite{method_orthoframes} propose a classification technique that utilizes ConvNets and ResNet101 to identify pavement distress. The method relies on orthoframes obtained from a mobile mapping system in Italy. Stricker \etal \cite{method_gaps_dataset} employ a pretrained ResNet34 model to enhance the evaluation of road conditions by utilizing transfer learning. The model underwent testing using privately acquired data from Germany, encompassing six distinct categories of road conditions. Hsieh \etal \cite{method_raveling} devised a classification model that utilizes transfer learning and ResNet18 to identify pavement raveling. The model underwent testing using a proprietary dataset obtained within the borders of the United States. Sajad \etal \cite{method_bleeding} devised a classification model for inspecting pavement bleeding using transfer learning. The methodology was evaluated using a proprietary dataset acquired in Iran. Gopalakrishnan \etal \cite{method_cnn_2017} devised a classification system based on Deep Neural Networks, specifically utilizing the VGG-16 model. The purpose of this system was to accurately detect and classify pavement cracks in a dataset obtained from the United States. Nguyen \etal \cite{method_cnn_soict} and Nhat \etal \cite{method_crack_pave_vn} developed a CNN structure for the purpose of identifying pavement cracks on roads in Vietnam. Li \etal \cite{method_crack_pave_china} developed a specialized CNN model for identifying pavement cracks in China. In addition, with the rapid growth of 3D vision such as \cite{mdec_2023}, lidar-based 3D cracks of road surface has been well studied such as \cite{feng2021gcn_lidar_fusion}.

In summary, transfer learning is a common approach for developing classification models that adapt to road surface conditions detection and perform well across a wide range of road condition sets and locations. However, in road surface condition inspection based on images, using the entire image as described above could be sufficient because the model can learn from any visual information in the image that is not directly related to the road. In the following section, we present our method for improving the classification model based on the identified drawback.

\section{Method} \label{sec:method}
The workflow of our proposed framework is represented in Figure \ref{fig:our_fw}. In our framework, we extract road area from the original image by first converting it to a semantic map with road masking, then extracting the road area mask from the segmentation map, and then extracting the road area from the original image. The extracted road area is then fed into the road surface condition classification model to determine the state of the road surface, \ie, $C_1,C_2,...,C_n$.  The technical details of our framework are described below. 

\subsection{Road area extraction}
In this stage, we first train an encoder-decoder model with a segmentation task that is separate from the classification model trained in the second stage to predict a segmentation map from an image. Our aim is to segment a given image into two parts: road area and non-road area, therefore this is a segmentation task with binary classes. We proceed by resizing an image to a resolution of $H\times W\times C$, where $H,W,C$ represent the image's height, width, and number of channels. The model generates a segmentation map of size $H\times W$, with each pixel belonging to one of two classes: $1$ for road areas and $0$ for non-road areas like cars, buildings, and backgrounds. We train the segmentation model using binary cross entropy loss function $\mathcal{L}_{seg}$ as shown below:
\begin{equation}
    \mathcal{L}_{seg} = -\frac{1}{N}\sum_{i=1}^N y_i\log(P_i)+(1-y_i)\log(1-P_i)
\end{equation}
where $N$ is the number of pixels in a training batch, $y_i$ and $P_i$ are annotated and predicted confidence of a pixel, respectively.
Once we have trained the segmentation model, we use it to extract the segmentation map from given images. From the segmentation map, we create a segmented road mask in which only road type pixels are maintained and non-road pixels are set to 0. The segmented road map is then used to extract the road area from the resized original image. The extracted road area from the original image is more useful for the classification model than a segmented road mask. Finally, the image containing only the road region is resized to fit onto the classification model created in the next phase.
\subsection{Classification model}
\begin{figure}[!ht]
    \centering
    \includegraphics[width=0.49\textwidth]{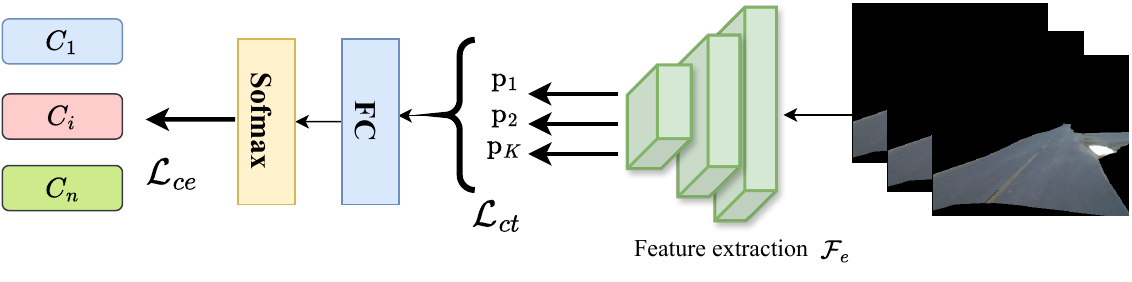}
    \caption{ Classification model for road surface conditions with contrastive learning to enhance performance of classification task.}
    \label{fig:classification}
\end{figure}
We construct a classification model for road surface condition classification. We use the extracted road area as input for classification models. The classification model takes input as a road area and aims to classify it into $n$ classes, ranging from $C_1$ to $C_n$. At this stage, the extracted road area are sent into a feature extraction network $\mathcal{F}_e$ to extract semantic embedding features. We have used a pretrained backbone as $\mathcal{F}_e$ for enhanced feature extraction.

Embedding features are important for classification model. In cases in which there is a significant lack of consistency within a class, the semantic embedding features may exhibit heterogeneity. Therefore, in this work, we have employed contrastive learning \cite{contrastive,contrastive2,dg2hairoad} to improve the classification task. Given a pair of samples ($\mathrm{x}_i, \mathrm{x}_j$), we denote the corresponding extracted semantic embedding feature as $\mathrm{p}_i$ and $\mathrm{p}_j$, respectively. During training, we apply contrastive loss to each positive pair ($\mathrm{x}_i, \mathrm{x}_j$) in each training batch of $K$ data, as shown in the equation below:
\begin{equation} \label{eq:consLoss}
\mathcal{L}_{ct}(\mathrm{x}_i,\mathrm{x}_j)=-\log{\frac{\exp(\frac{SIM(\mathrm{x}_i,\mathrm{x}_j)}{\tau})}{\sum_{k=1}^{K}I(\mathrm{x}_i,\mathrm{x}_k)\cdot \exp(\frac{SIM(\mathrm{x}_i,\mathrm{x}_j)}{\tau})}}
\end{equation}
where $I$ denote an indicator function as below equation:
\begin{equation} \label{eq:identity}
 I(\mathrm{x}_i,\mathrm{x}_k) =
    \begin{cases}
      0 & \mathrm{x}_i,\mathrm{x}_k \text{ is same class}\\
      1 & \text{otherwise}
    \end{cases}       
\end{equation}
Instead of calculating the similarity of two original images $\mathrm{x}_i$ and $\mathrm{x}_j$ directly, we compute this similarity using the corresponding embedded feature. For further details, we use the cosine distance between two features to determine the similarity of two vectors. The similarity of a pair $(\mathrm{x}_i,\mathrm{x}_j)$ is determined using the equation below:
\begin{equation} \label{eq:similarity}
 SIM(\mathrm{x}_i,\mathrm{x}_j)\approx cosine(\mathrm{p}_i,\mathrm{p}_j)=\frac{\mathrm{p}_i^T\times \mathrm{p}_j}{||\mathrm{p}_i||\times||\mathrm{p}_j||}
\end{equation}
The contrastive loss function is calculated for all positive pairs in the batch training for both $(\mathrm{x}_i,\mathrm{x}_j)$ and $(\mathrm{x}_j,\mathrm{x}_i)$. Training in such a way improves the capacity of model to explore the invariance of representations, allowing embedding features of the same class to be close to one another while keeping those of different classes apart.

Finally, the training loss of our model is described below:
\begin{equation} \label{eq:trainLoss}
\mathcal{L}=\mathcal{L}_{ce} + \lambda\mathcal{L}_{ct}
\end{equation}
where $\lambda$ is a hyperparameter for weight of contrastive loss, and $\mathcal{L}_{ce}$ is the categorical cross-entropy loss \cite{categorical_cross_entropy}.

\section{\uppercase{Experiment}} \label{sec:exp}
\begin{table*}[!ht]
\centering
\caption{Results of road condition classification with a comparison to the baseline and our method using same classifiers.}
\label{table:result}
\resizebox{\textwidth}{!}{%
\begin{tabular}{|l||cccccc||cccccc||cccccc||c|}
\hline
\multicolumn{1}{|c||}{\multirow{3}{*}{\textbf{Class}}} & \multicolumn{6}{c||}{\textbf{ResNet18-backbone}}                                                                                                                                                               & \multicolumn{6}{c||}{\textbf{VGG16-backbone}}                                                                                                                                                            & \multicolumn{6}{c||}{\textbf{MobileNetV2-backbone}}                                                                                                                                                      & \multirow{3}{*}{\textbf{Supports}} \\ \cline{2-19}
\multicolumn{1}{|c||}{}                                & \multicolumn{3}{c|}{\textbf{Hsieh \cite{method_raveling}}}                                                                     & \multicolumn{3}{c||}{\textbf{Ours.}}                                                     & \multicolumn{3}{c|}{\textbf{Hasanaath \cite{method_transfer_learning}}}                                                               & \multicolumn{3}{c||}{\textbf{Ours.}}                                                     & \multicolumn{3}{c|}{\textbf{Chaudhary \cite{method_light_weight_nn}}}                                                               & \multicolumn{3}{c||}{\textbf{Ours.}}                                                     &                                    \\ \cline{2-19}
\multicolumn{1}{|c||}{}                                & \multicolumn{1}{c|}{\textbf{P}}    & \multicolumn{1}{c|}{\textbf{R}}    & \multicolumn{1}{c|}{\textbf{F1}} & \multicolumn{1}{c|}{\textbf{P}}    & \multicolumn{1}{c|}{\textbf{R}}    & \textbf{F1}   & \multicolumn{1}{c|}{\textbf{P}} & \multicolumn{1}{c|}{\textbf{R}} & \multicolumn{1}{c|}{\textbf{F1}} & \multicolumn{1}{c|}{\textbf{P}}    & \multicolumn{1}{c|}{\textbf{R}}    & \textbf{F1}   & \multicolumn{1}{c|}{\textbf{P}} & \multicolumn{1}{c|}{\textbf{R}} & \multicolumn{1}{c|}{\textbf{F1}} & \multicolumn{1}{c|}{\textbf{P}}    & \multicolumn{1}{c|}{\textbf{R}}    & \textbf{F1}   &                                    \\ \hline
asphalt bad                                           & \multicolumn{1}{c|}{0.82}          & \multicolumn{1}{c|}{0.92}          & \multicolumn{1}{c|}{0.87}        & \multicolumn{1}{c|}{\textbf{0.88}} & \multicolumn{1}{c|}{\textbf{0.95}} & \textbf{0.91} & \multicolumn{1}{c|}{0.82}       & \multicolumn{1}{c|}{0.83}       & \multicolumn{1}{c|}{0.82}        & \multicolumn{1}{c|}{\textbf{0.89}} & \multicolumn{1}{c|}{\textbf{0.88}} & \textbf{0.89} & \multicolumn{1}{c|}{0.91}       & \multicolumn{1}{c|}{0.86}       & \multicolumn{1}{c|}{0.88}        & \multicolumn{1}{c|}{\textbf{0.98}} & \multicolumn{1}{c|}{\textbf{0.92}} & \textbf{0.95} & 316                                \\ \hline
asphalt good                                          & \multicolumn{1}{c|}{\textbf{0.96}} & \multicolumn{1}{c|}{\textbf{0.99}} & \multicolumn{1}{c|}{0.97}        & \multicolumn{1}{c|}{\textbf{0.96}} & \multicolumn{1}{c|}{\textbf{0.99}} & \textbf{0.98} & \multicolumn{1}{c|}{0.93}       & \multicolumn{1}{c|}{0.97}       & \multicolumn{1}{c|}{0.95}        & \multicolumn{1}{c|}{\textbf{0.95}} & \multicolumn{1}{c|}{\textbf{0.99}} & \textbf{0.97} & \multicolumn{1}{c|}{0.98}       & \multicolumn{1}{c|}{0.97}       & \multicolumn{1}{c|}{0.97}        & \multicolumn{1}{c|}{\textbf{0.99}} & \multicolumn{1}{c|}{\textbf{0.99}} & \textbf{0.99} & 1371                               \\ \hline
asphalt regular                                       & \multicolumn{1}{c|}{0.94}          & \multicolumn{1}{c|}{0.86}          & \multicolumn{1}{c|}{0.90}        & \multicolumn{1}{c|}{\textbf{0.97}} & \multicolumn{1}{c|}{\textbf{0.89}} & \textbf{0.93} & \multicolumn{1}{c|}{0.85}       & \multicolumn{1}{c|}{0.84}       & \multicolumn{1}{c|}{0.84}        & \multicolumn{1}{c|}{\textbf{0.9}}  & \multicolumn{1}{c|}{\textbf{0.89}} & \textbf{0.89} & \multicolumn{1}{c|}{0.91}       & \multicolumn{1}{c|}{0.92}       & \multicolumn{1}{c|}{0.91}        & \multicolumn{1}{c|}{\textbf{0.94}} & \multicolumn{1}{c|}{\textbf{0.95}} & \textbf{0.94} & 593                                \\ \hline
concrete bad                                          & \multicolumn{1}{c|}{0.76}          & \multicolumn{1}{c|}{0.81}          & \multicolumn{1}{c|}{0.78}        & \multicolumn{1}{c|}{\textbf{0.88}} & \multicolumn{1}{c|}{\textbf{0.91}} & \textbf{0.89} & \multicolumn{1}{c|}{0.59}       & \multicolumn{1}{c|}{0.58}       & \multicolumn{1}{c|}{0.58}        & \multicolumn{1}{c|}{\textbf{0.69}} & \multicolumn{1}{c|}{\textbf{0.59}} & \textbf{0.64} & \multicolumn{1}{c|}{0.72}       & \multicolumn{1}{c|}{0.83}       & \multicolumn{1}{c|}{0.77}        & \multicolumn{1}{c|}{\textbf{0.86}} & \multicolumn{1}{c|}{\textbf{0.97}} & \textbf{0.91} & 88                                 \\ \hline
concrete regular                                      & \multicolumn{1}{c|}{0.94}          & \multicolumn{1}{c|}{0.83}          & \multicolumn{1}{c|}{0.88}        & \multicolumn{1}{c|}{\textbf{0.98}} & \multicolumn{1}{c|}{\textbf{0.88}} & \textbf{0.93} & \multicolumn{1}{c|}{0.75}       & \multicolumn{1}{c|}{0.61}       & \multicolumn{1}{c|}{0.67}        & \multicolumn{1}{c|}{\textbf{0.8}}  & \multicolumn{1}{c|}{\textbf{0.67}} & \textbf{0.73} & \multicolumn{1}{c|}{0.83}       & \multicolumn{1}{c|}{0.85}       & \multicolumn{1}{c|}{0.84}        & \multicolumn{1}{c|}{\textbf{0.93}} & \multicolumn{1}{c|}{\textbf{0.94}} & \textbf{0.93} & 233                                \\ \hline
unpaved bad                                           & \multicolumn{1}{c|}{0.93}          & \multicolumn{1}{c|}{0.96}          & \multicolumn{1}{c|}{0.94}        & \multicolumn{1}{c|}{\textbf{0.95}} & \multicolumn{1}{c|}{\textbf{1}}    & \textbf{0.97} & \multicolumn{1}{c|}{0.87}       & \multicolumn{1}{c|}{0.95}       & \multicolumn{1}{c|}{0.91}        & \multicolumn{1}{c|}{\textbf{0.91}} & \multicolumn{1}{c|}{\textbf{0.97}} & \textbf{0.94} & \multicolumn{1}{c|}{0.95}       & \multicolumn{1}{c|}{\textbf{1}} & \multicolumn{1}{c|}{0.97}        & \multicolumn{1}{c|}{\textbf{0.97}} & \multicolumn{1}{c|}{\textbf{1}}    & \textbf{0.98} & 418                                \\ \hline
unpaved regular                                       & \multicolumn{1}{c|}{0.98}          & \multicolumn{1}{c|}{0.94}          & \multicolumn{1}{c|}{0.96}        & \multicolumn{1}{c|}{\textbf{1}}    & \multicolumn{1}{c|}{\textbf{0.97}} & \textbf{0.98} & \multicolumn{1}{c|}{0.97}       & \multicolumn{1}{c|}{0.92}       & \multicolumn{1}{c|}{0.94}        & \multicolumn{1}{c|}{\textbf{0.99}} & \multicolumn{1}{c|}{\textbf{0.95}} & \textbf{0.97} & \multicolumn{1}{c|}{0.99}       & \multicolumn{1}{c|}{0.94}       & \multicolumn{1}{c|}{0.96}        & \multicolumn{1}{c|}{\textbf{1}}    & \multicolumn{1}{c|}{\textbf{0.98}} & \textbf{0.99} & 565                                \\ \hline
\textbf{Macro avg}                                    & \multicolumn{1}{c|}{0.90}          & \multicolumn{1}{c|}{0.90}          & \multicolumn{1}{c|}{0.90}        & \multicolumn{1}{c|}{\textbf{0.95}} & \multicolumn{1}{c|}{\textbf{0.94}} & \textbf{0.94} & \multicolumn{1}{c|}{0.83}       & \multicolumn{1}{c|}{0.81}       & \multicolumn{1}{c|}{0.82}        & \multicolumn{1}{c|}{\textbf{0.88}} & \multicolumn{1}{c|}{\textbf{0.85}} & \textbf{0.86} & \multicolumn{1}{c|}{0.90}       & \multicolumn{1}{c|}{0.91}       & \multicolumn{1}{c|}{0.90}        & \multicolumn{1}{c|}{\textbf{0.95}} & \multicolumn{1}{c|}{\textbf{0.94}} & \textbf{0.96} & 3584                               \\ \hline
\end{tabular}
}
\end{table*}
\subsection{Datasets}
For our experiment, we use CityScapes \cite{cityscape} to train the segmentation model in the initial phase. For binary segmentation, we pre-process the annotation of CityScapes dataset in which we keep only road class, while the other classes are merged as background. To provide further elaboration, we conducted segmentation training using a dataset consisting of 2,975 samples, and other 500 samples were reserved for validation purposes and to determine the best checkpoint. The segmentation model takes an input image of size 352x288 and generates a segmentation output of the same size.

In this experiment, we simulate a road surface condition classification task using multiple types of road surfaces, each with its own set of distinct measurement conditions. To support our experiment, we use RTK classification dataset \cite{rtk_road_quality}. This dataset includes three road surface types: asphalt, concrete, and unpaved. The conditions are good, regular, and bad for asphalt, as well as regular and bad for concrete and unpaved roads. Overall, there are seven types of road conditions to determine: asphalt bad, asphalt good, asphalt regular, concrete bad, concrete regular, unpaved bad, and unpaved regular. We divided the whole dataset of 5,118 images into three sets: training, validation, and test, with ratios of 0.2, 0.1, and 0.7, respectively. The training, validation, and test sets each have a sample size of \{101, 399, 161, 28, 63, 112, 159\}, \{47, 208, 85, 8, 28, 63, 72\}, and \{316, 1371, 593, 88, 233, 418, 565\} respectively. During training, we use the validation set to find the best checkpoint. The test set is used to produce the results for our report. We resize images to 128x128, and normalize them for both training/validation and testing sets. We use horizontal flipping for data augmentation only on the training dataset.

\subsection{Settings}
To compare to previous work, we chose three recent models for road condition classification: Hsieh \cite{method_raveling}, Hasanaath \cite{method_transfer_learning}, Chaudhary \cite{method_light_weight_nn}. These models use ImageNet's pretrained backbone for feature extraction. For our method, we use feature extraction $\mathcal{F}_e$ of classification model as one of these pretrained backbones. For segmentation model, we train several architectures which are FPN \cite{fpn_seg}, PSPNet \cite{pspnet}, and UNet \cite{unet}, UNet++ \cite{unet++} on CityScapes dataset for road extraction further.
We train the segmentation and classification models in a batch size of 32 in 20 epochs using Adam optimizer with a learning rate of $1e^{-4}$.
For our classification models, we set $\lambda$ to 1.0 and temperature parameter $\tau$ to 0.05. 
All experiments were carried out on an NVIDIA GeForce RTX 4090 23GB GPU with 32GB of RAM, using the PyTorch library.

\textbf{Evaluation metrics.} We use precision, recall and F1-score for evaluating classification models.
\subsection{Results}
Table \ref{table:result} compares our method to three recent models: Hsieh \cite{method_raveling}, Hasanaath \cite{method_transfer_learning}, Chaudhary \cite{method_light_weight_nn}. The results show that all models are more effective at distinguishing three road conditions classes: asphalt good, unpaved bad, and unpaved regular than other road conditions. However, concrete bad condition is more confusing than other classes and more difficult to identify. Concrete surface defects, such as dirt and water, may lead to noisy features which contribute misclassification. When compared to recent approaches Hsieh \cite{method_raveling}, Hasanaath \cite{method_transfer_learning}, Chaudhary \cite{method_light_weight_nn} has lower classification performance across all classes. Our method improved the performance of all classifiers over their baseline. The results show that our method significantly improves precision, recall, and F1-score across all road condition classes. It indicates that our method is promising for classifying road conditions using images.

\textbf{Impact of segmentation models.} We conduct an analysis of the impact of the segmentation model on the performance of our model. We train several state-of-the-art segmentation model architectures in Cityscapes, including FPN \cite{fpn_seg}, PSPNet \cite{pspnet}, and UNet \cite{unet}. Then we apply these models as the segmentation model in our framework. Table \ref{table:seg_arch} shows the results of road condition classification using various segmentation models in our framework. Here we use ResNet18 in the classification phase. The results show that UNet++ enhances road condition classification performance compared to other segmentation models. It is reasonable because UNet++ outperformed other segmentation models in Cityscapes validation, as shown in Table \ref{table:seg_arch}. The performance on the CiteScape dataset with observed DICE metric in this table indicates that a better segmentation model is more appropriate for better classification of the entire framework. It indicates that selecting the appropriate segmentation has a positive impact on the performance of road condition classification.
\begin{table}[ht]
\centering
\caption{Performance of road condition classification of our model using on variety of segmentation model architectures. Segmentation models were also evaluated in the CityScapes validation set with DICE score.}
\label{table:seg_arch}
\begin{tabular}{|l|c|ccc|}
\hline
\multirow{2}{*}{\textbf{Archs}} & \textbf{CityScapes} & \multicolumn{3}{c|}{\textbf{Road condition classification}}                                                                  \\ \cline{2-5} 
                                & \textbf{DICE}       & \multicolumn{1}{c|}{\textbf{Precision}} & \multicolumn{1}{c|}{\textbf{Recall}} & \textbf{F1-score} \\ \hline
FPN                             & 0.476               & \multicolumn{1}{c|}{0.91}               & \multicolumn{1}{c|}{0.9}             & 0.90              \\ \hline
PSPNet                          & 0.503               & \multicolumn{1}{c|}{0.93}               & \multicolumn{1}{c|}{0.92}            & 0.92              \\ \hline
UNet                            & 0.482               & \multicolumn{1}{c|}{0.92}               & \multicolumn{1}{c|}{0.9}             & 0.91              \\ \hline
UNet++                         & 0.533               & \multicolumn{1}{c|}{0.95}               & \multicolumn{1}{c|}{0.94}            & 0.94              \\ \hline
\end{tabular}
\end{table}

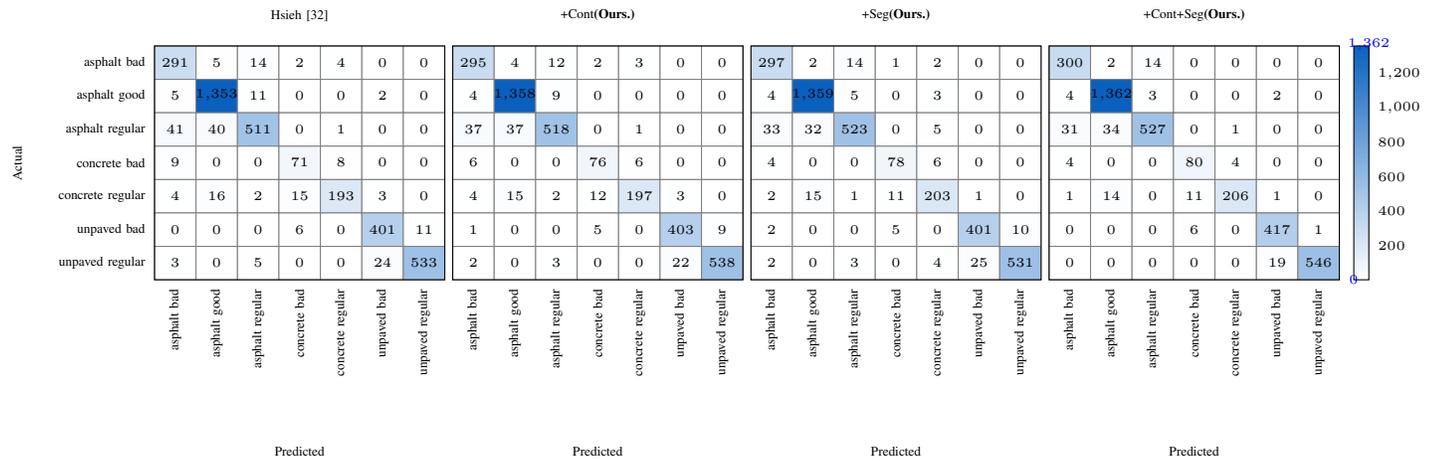
\begin{figure*}[!ht]
\centering
\begin{tikzpicture}[font=\tiny]

    \begin{groupplot}[%
            group style={
                columns=4,
                group name=plots,
                x descriptions at=edge bottom,
                y descriptions at=edge left,
                vertical sep=0pt,
                horizontal sep=3pt
            },
            width=.3\textwidth,
            colormap={bluewhite}{color=(white) rgb255=(9,96,191)},
            group/xlabels at = edge bottom,
            xlabel=Predicted,
            xlabel style={yshift=-20pt},
            ylabel=Actual,
            ylabel style={yshift=7pt},
            xticklabels={asphalt bad, asphalt good, asphalt regular, concrete bad, concrete regular, unpaved bad, unpaved regular}, 
            xtick={0,...,6}, 
            xtick style={draw=none},
            yticklabels={asphalt bad, asphalt good, asphalt regular, concrete bad, concrete regular, unpaved bad, unpaved regular}, 
            ytick={0,...,6}, 
            ytick style={draw=none},
            enlargelimits=false,
            xticklabel style={
              rotate=90
            },
            nodes near coords={\pgfmathprintnumber\pgfplotspointmeta},
            nodes near coords style={
                yshift=-5pt
            }
        ]
        \nextgroupplot[title=Hsieh \cite{method_raveling}]
        \addplot[
            matrix plot,
            mesh/cols=7, 
            point meta=explicit,draw=gray
        ] table [meta=C] {
            x y C
            0 0 291
            1 0 5
            2 0 14
            3 0 2
            4 0 4
            5 0 0
            6 0 0
            
            0 1 5
            1 1 1353
            2 1 11
            3 1 0
            4 1 0
            5 1 2
            6 1 0
            
            0 2 41
            1 2 40
            2 2 511
            3 2 0
            4 2 1
            5 2 0
            6 2 0
    
            0 3 9
            1 3 0
            2 3 0
            3 3 71
            4 3 8
            5 3 0
            6 3 0

            0 4 4
            1 4 16
            2 4 2
            3 4 15
            4 4 193
            5 4 3
            6 4 0

            0 5 0
            1 5 0
            2 5 0
            3 5 6
            4 5 0
            5 5 401
            6 5 11

            0 6 3
            1 6 0
            2 6 5
            3 6 0
            4 6 0
            5 6 24
            6 6 533
        
        }; 

        \nextgroupplot[title=\text{+Cont}\textbf{(Ours.)}]
        \addplot[
            matrix plot,
            mesh/cols=7, 
            point meta=explicit,draw=gray
        ] table [meta=C] {
            x y C
            0 0 295
            1 0 4
            2 0 12
            3 0 2
            4 0 3
            5 0 0
            6 0 0
            
            0 1 4
            1 1 1358
            2 1 9
            3 1 0
            4 1 0
            5 1 0
            6 1 0
            
            0 2 37
            1 2 37
            2 2 518
            3 2 0
            4 2 1
            5 2 0
            6 2 0
    
            0 3 6
            1 3 0
            2 3 0
            3 3 76
            4 3 6
            5 3 0
            6 3 0

            0 4 4
            1 4 15
            2 4 2
            3 4 12
            4 4 197
            5 4 3
            6 4 0

            0 5 1
            1 5 0
            2 5 0
            3 5 5
            4 5 0
            5 5 403
            6 5 9

            0 6 2
            1 6 0
            2 6 3
            3 6 0
            4 6 0
            5 6 22
            6 6 538
        
        }; 
        \nextgroupplot[title=\text{+Seg}\textbf{(Ours.)}]
        \addplot[
            matrix plot,
            mesh/cols=7, 
            point meta=explicit,draw=gray
        ] table [meta=C] {
            x y C
            0 0 297
            1 0 2
            2 0 14
            3 0 1
            4 0 2
            5 0 0
            6 0 0
            
            0 1 4
            1 1 1359
            2 1 5
            3 1 0
            4 1 3
            5 1 0
            6 1 0
            
            0 2 33
            1 2 32
            2 2 523
            3 2 0
            4 2 5
            5 2 0
            6 2 0
    
            0 3 4
            1 3 0
            2 3 0
            3 3 78
            4 3 6
            5 3 0
            6 3 0

            0 4 2
            1 4 15
            2 4 1
            3 4 11
            4 4 203
            5 4 1
            6 4 0

            0 5 2
            1 5 0
            2 5 0
            3 5 5
            4 5 0
            5 5 401
            6 5 10

            0 6 2
            1 6 0
            2 6 3
            3 6 0
            4 6 4
            5 6 25
            6 6 531
        
        }; 
        
        \nextgroupplot[
            title=\text{+Cont+Seg}\textbf{(Ours.)},
            colorbar right,
            colorbar style= {
                at={(rel axis cs: 4.08,0)},
                anchor=south,
                width=2mm,
                ytick={200,400,600,800,1000,1200,1400}, 
            }
            ]
        \addplot[
            matrix plot,
            mesh/cols=7, 
            point meta=explicit,draw=gray
        ] table [meta=C] {
            x y C
            0 0 300
            1 0 2
            2 0 14
            3 0 0
            4 0 0
            5 0 0
            6 0 0
            
            0 1 4
            1 1 1362
            2 1 3
            3 1 0
            4 1 0
            5 1 2
            6 1 0
            
            0 2 31
            1 2 34
            2 2 527
            3 2 0
            4 2 1
            5 2 0
            6 2 0
    
            0 3 4
            1 3 0
            2 3 0
            3 3 80
            4 3 4
            5 3 0
            6 3 0

            0 4 1
            1 4 14
            2 4 0
            3 4 11
            4 4 206
            5 4 1
            6 4 0

            0 5 0
            1 5 0
            2 5 0
            3 5 6
            4 5 0
            5 5 417
            6 5 1

            0 6 0
            1 6 0
            2 6 0
            3 6 0
            4 6 0
            5 6 19
            6 6 546
        
        }; 
        
    \end{groupplot}
\end{tikzpicture}
\caption{Ablation study with baseline is Hsieh \cite{method_raveling}. '+Cont(\textbf{Ours.})': training baseline classifier with our incorporated contrastive learning, '+Seg(\textbf{Ours.})': using our extracted road area with segmentation for training and testing with baseline classifier.}
\label{fig:ablation_study}
\end{figure*}
\textbf{Ablation study.} To further investigate the impact of each component in our method, we conduct an ablation study using $\mathcal{F}_e$ as a ResNet18 pretrained on ImageNet. Figure \ref{fig:ablation_study} shows the results of the ablation study in the confusion matrix of prediction and ground truth. Our method improves prediction of all road condition classes by training a baseline classifier employing constrastive learning. It also reduces the number of significant missing predictions for each class.  It demonstrates that contrastive learning is effective at distinguishing confusing classes. For the next component, training and testing baseline models with road area extraction using our method improves the overall accuracy of road condition classes. However, compared to contrastive learning, the improvement in unpaved bad and unpaved regular is lower. It is reasonable because our segmentation model was trained on the CityScapes dataset, which only included roads in cities. The segmentation model may not perform well on unseen, unpaved roads, resulting in lower performance than contrastive learning. On the other hand, the improvement of remaining classes was better with contrastive learning. It indicates that extracting road area is significantly important for improving road condition classification performance. Finally, our method, which uses extracted road area as training and testing inputs and incorporates contrastive learning for training, improves the classification model's performance significantly across all classes.  

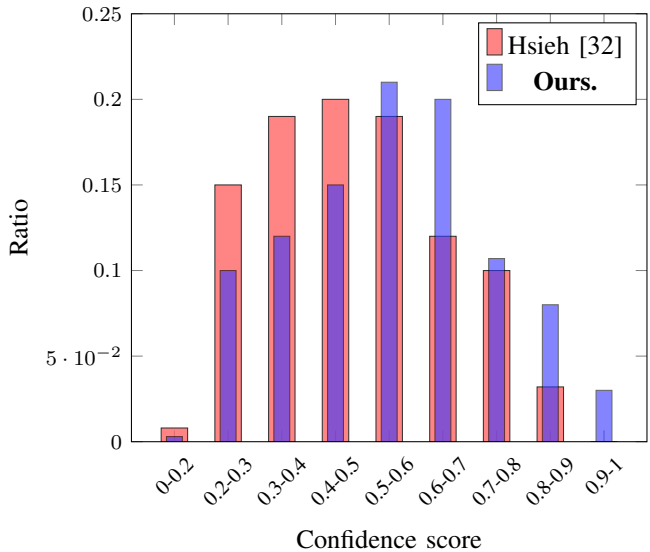
\begin{figure}[!ht]
\centering
    \begin{tikzpicture}
    \begin{axis}[
      ymin=0,
      ymax=0.25,
      symbolic x coords={0-0.2,0.2-0.3,0.3-0.4,0.4-0.5,0.5-0.6,0.6-0.7,0.7-0.8,0.8-0.9,0.9-1},
      xtick={0-0.2,0.2-0.3,0.3-0.4,0.4-0.5,0.5-0.6,0.6-0.7,0.7-0.8,0.8-0.9,0.9-1},
      legend image code/.code={
        \draw [#1] (0cm,-0.1cm) rectangle (0.2cm,0.25cm); },
        ylabel=Ratio,
        yticklabel style={font=\footnotesize},
        xticklabel style={rotate=45,font=\footnotesize},
        xlabel={Confidence score},
        xlabel near ticks,
    ]
    \addplot[ybar,bar width=10pt,fill=red!60,opacity=0.8] 
      table[col sep=comma,y index=1] {data1.txt};
      
    \addplot[ybar,bar width=6pt,fill=blue!80,opacity=0.6] 
      table[col sep=comma,y index=2] {data1.txt};
      
    \legend{Hsieh \cite{method_raveling},\textbf{Ours.}}
    \end{axis}
    \end{tikzpicture}
\caption{Comparison of our method with model Hsieh \cite{method_raveling} in terms of ratio of confidence score of predicted road condition class.}
\label{fig:ratio_conf_score}
\end{figure}
\textbf{Confidence of model classification.} In Figure \ref{fig:ratio_conf_score}, we compare our method to model Hsieh \cite{method_raveling} in terms of predicted class confidence score ratio. The confidence score is the maximum probability derived from model prediction that is used to determine the output class. We divide the confidence score range into nine groups and then calculate the ratio of each range using the predicted class's confidence score. The Figure \ref{fig:ratio_conf_score} shows that our method predicts with a higher ratio of confidence score than Hsieh \cite{method_raveling} between the ranges of 0.5-0.6 and 0.9-1. This result shows that our method significantly improves the confidence of model classification, which is essential to improve classification performance for road conditions.

\section{Conclusions}\label{sec:conclude}
This paper presents an improved method for classifying road surface conditions. We present a road area extraction module employing a segmentation model as input to the model. Furthermore, we propose using contrastive learning to improve classification performance. The results of our research on the RTK dataset with seven road surface condition classes demonstrates that our model outperforms certain classifiers' baseline models significantly. It demonstrates our method's effective and critical positive impact on road condition classification. In the future, we intend to expand our method to other tasks such as road condition detection and multiple class road condition classification.

\section*{Acknowledgment}
This work was realized in imec.ICON Hybrid AI for predictive road maintenance (HAIROAD) project, with the financial support of Flanders Innovation \& Entrepreneurship (VLAIO, project no. HBC.2023.0170).

\bibliographystyle{IEEEtran}
\bibliography{reference.bib}

\end{document}